\documentclass{article}

\usepackage{aimc2022}

\usepackage[utf8]{inputenc} 
\usepackage[T1]{fontenc}    
\usepackage{hyperref}       
\usepackage{url}            
\usepackage{booktabs}       
\usepackage{amsfonts}       
\usepackage{nicefrac}       
\usepackage{microtype}      
\usepackage{graphicx}       
\usepackage[usenames,dvipsnames]{xcolor}
\newcommand{\linkcolor}{MidnightBlue}
\hypersetup{
  colorlinks=true,
  citecolor=\linkcolor,
  linkcolor=\linkcolor,
  urlcolor=\linkcolor,
  citebordercolor=\linkcolor,
  linkbordercolor=\linkcolor,
  urlbordercolor=\linkcolor,
  pdfborderstyle={/S/U/W 1}
}
\makeatletter
\Hy@AtBeginDocument{%
  \def\@pdfborder{0 0 1}
  \def\@pdfborderstyle{/S/U/W 0.5}
}
\makeatother

\newcommand{\smurl}{http://aimc2022-suppl-mat.s3-website.eu-west-3.amazonaws.com/index.html}

\title{``Melatonin'': A Case Study on AI-induced\\ Musical Style}

\author{%
Emmanuel Deruty
\\
Sony Computer Science Laboratories (CSL) Paris\\
6 Rue Amyot\\
75005 Paris, France\\
\url{https://csl.sony.fr/team/emmanuel-deruty/}\\
\And
Maarten Grachten \\
Contractor for Sony CSL Paris\\
6 Rue Amyot\\
75005 Paris, France\\
\url{https://maarten.grachten.eu} \\
}

\begin{document}

\maketitle

\begin{abstract}
  Although the use of AI tools in music composition and production is steadily increasing, as witnessed by the newly founded AI song contest, analysis of music produced using these tools is still relatively uncommon as a mean to gain insight in the ways AI tools impact music production.
  In this paper we present a case study of ``Melatonin'', a song produced by extensive use of BassNet, an AI tool originally designed to generate bass lines.
  Through analysis of the artists' work flow and song project, we identify style characteristics of the song in relation to the affordances of the tool, highlighting manifestations of style in terms of both idiom and sound.
\end{abstract}

\section{Introduction and Related Work}\label{sec:introduction}

AI is rapidly gaining foothold in the workflow of music artists, bringing exciting opportunities for musical creativity.
Music AI tools offer novel functionality and affordances \citep{mcgrenere00:_affor}, and a particularly interesting question in this context is whether, and how, this novelty affects stylistic aspects of music made using these tools.

Obviously it is impossible to provide a general response to this question, not least because music AI tools form a very broad category that is only just emerging, but also because there are no established practices yet for the use of AI tools in music.
Nevertheless we believe it is instructive to study how individual music artists use music AI tools, and how these shape the resulting music, as it can further our understanding of what it is that allows AI tools to bring value to the music production workflow, and thereby help to improve the design and effectiveness of the tools.

We intend to contribute to this aim by way of a case study: we focus on a collaboration between music artist duo Hyper Music\footnote{\url{https://www.hyper-music.com/}} and the music research group of Sony Computer Science Laboratories (CSL) Paris\footnote{\url{https://cslmusicteam.sony.fr/}}, in which Hyper Music uses a CSL-developed AI tool to produce new songs.
We focus particularly on stylistic aspects of the resulting music, drawing on definitions from \cite{OUP_Style}.

This focus on the musical output itself, rather than feedback from the artists, makes our work complementary to works like \citep{huang2020ai} and \citep{deruty2022development}, where thematic analyses are made based on feedback from the artists/users reporting on their experience using the tools.
Other works evaluate music AI tools by qualitative judgments of value by humans.
For example, \cite{cadiz21:_creat_gener_music_networ} ask human subjects to rate the creativity of generated chords and MIDI piano excerpts.
\cite{collins_laney_willis_garthwaite_2016} report on a listening experiment in which subjects rate the degree to which the outputs of generative computational models of stylistic composition adhere to the characteristics of the intended style (Chopin's Mazurkas).
\cite{DBLP:journals/corr/SturmSBK16} evaluate a model of Celtic folk melodies by comparing the outputs to the training data in terms of statistical properties, by qualitative evaluation of the stylistic features of the outputs, and through the utility of the model as a compositional tool.
The above works focus on style in the context of symbolic representations of music.
As for style in the context of audio, existing work deals predominantly with the problem of \emph{style transfer}, e.g. transformation between music instruments \cite{10.1007/978-3-030-29891-3_29},
or morphing between sounds such as speech and music \cite{8461711}.

Our contributions in this paper are the following.
Firstly, we document the usage of a music AI tool by professional artists, as well as changes made to the tool in response to the artist's suggestions.
Secondly, we provide an analysis of a selection of the musical outputs in terms of their stylistic characteristics, establishing a link between these characteristics and the affordances of the AI tool.

The structure of the paper is as follows.
In Section~\ref{sec:background} we introduce the notion of musical style.
In Section~\ref{sec:method} we describe the form and the specific aims of the collaboration, as well as the methods of interpreting the musical outputs.
Section~\ref{sec:results} details how the AI tools were used in the music production process, and presents an analysis of selected musical outputs in terms of musical style.
We discuss the results in Section~\ref{sec:discussion}, and present our conclusions in Section~\ref{sec:conclusions}.

\section{Background: What is Musical Style?}\label{sec:background}
To address the question whether and how the AI tools used by Hyper Music affect the musical style of the outputs, we must first find an operational definition of the term.
According to the Merriam Webster dictionary, \emph{style} is defined as ``a distinctive manner of expression'' \citep{mw:style}.
In the field of music, style may qualify a variety of entities.
According to \citep{OUP_Style}, style refers to the manner in which a work of art is executed, and can denote music characteristic of an individual composer, a period, a geographical area or center, or of a society or social function.

For \cite{parry1911style}, musical style originates in the relationship between resources and utterance.
As such, style may also be linked to the resources or means by which the music is produced. 
This holds notably for the instruments used. Firstly, they may introduce characteristic \emph{sounds}, but by their specific affordances and limitations they may also lead to specific \emph{idioms} of expression.
Some of the examples \cite{OUP_Style} gives in this context are the violin with its capacity for wide-ranging melody and high tessitura, and the organ, where alternating feet on the bass pedals produce patterns that are distinct for late Baroque German organ music.
In contemporary Popular Music, tools used for generating and processing audio may play an equally prominent role in defining style.
Examples include the analog synthesizer, the distortion pedal, and more recently the ubiquitous Antares Auto-Tune plugin.

Following \citep{OUP_Style}, we distinguish between two manifestations of style, namely (1) the characteristic sounds stemming from the use of a given technology and (2) the idiomatic possibilities deriving from the technology.
We use these two aspects of style in our analysis of the outputs produced by Hyper Music.

\section{Data and Method}\label{sec:method}

\subsection{Collaboration between Hyper Music and CSL}\label{sec:hyper-music-sony}
\href{https://www.hyper-music.com/}{Hyper Music} is a sound and music production company that produces music for ads, TV series and feature films.
Over the years, Hyper Music has produced a catalog of music tracks.
When potential clients require music for an ad, they advertise their need and Hyper Music submits a track from their catalog.
If the track is selected for the ad, Hyper Music customizes the track for the client.
In that sense, Hyper Music is a provider of \textit{Production Music} \citep{Productionmusic}.

The \href{https://cslmusicteam.sony.fr/}{music research group} of Sony Computer Science Laboratories (CSL) Paris develops AI-based music tools to support music artists in their creative workflows.
To reinforce the practical relevance, development is done in parallel with collaborations in which professional music artists use prototypes of the tools.
Some of the collaborations have been reported by \cite{deruty2022development}.
The collaborations take place under an agreement where the artists agree to deliver one or more songs, making use of the AI tools provided by CSL.

An earlier collaboration of CSL with Hyper Music focused on embedding AI tools in their workflow with the aim of producing songs that would fit into the Hyper Music portfolio.
This collaboration was successful in its own right, in the sense that the music created with the help of CSL tools was used as the soundtrack in a worldwide \href{https://youtu.be/02_sciFL_XY}{Azzaro commercial}.
Nevertheless, the contribution of the AI tools in this material was subtle and restricted to interventions that did not noticeably alter the musical style---a consequence of the need of Hyper Music to maintain the trademark style of their portfolio.

To get a clearer view on the effect AI tools may have on musical style, a follow-up collaboration was planned, with a different specification of the task.
In this collaboration, the goal was to be more experimental and center the songs around the outputs of the AI tools, without the need for the resulting songs to fit into the portfolio.
This is not to say the aim was to produce \emph{experimental music}---a musical genre characterized by its radical opposition to and questioning of institutionalized modes of composition, performance, and aesthetics \citep{Experimentalmusic}.
On the contrary, the artists were asked not deviate from the codes of mainstream Popular Music they typically adhere to: there should be a stable tempo, an alternation between kick drum and snare drum, a reference to tonality, and a recognizable large-scale structure.

This follow-up collaboration took place over the course of 2021, and resulted in five tracks, using various CSL tools in the process. Most of these tools have been described by \cite{deruty2022development}.
In this paper we will focus on one of the five tracks -- titled ``Melatonin'',  which came about during a dedicated four-day collaborative session between Hyper Music and CSL, centered on CSL's \emph{BassNet} tool \citep{grachten2020bassnet}.
During this session CSL research scientists assisted Hyper Music with the use of the tool, observed the production strategies Hyper Music employed, and made changes to the BassNet prototype in response to suggestions by Hyper Music.

\subsection{BassNet}
\label{sec:bassnet}
BassNet is a music AI tool by CSL, designed to create bass lines to accompany existing audio material provided by the user (we refer to the latter as \emph{conditioning material/audio}). It takes any mix of existing audio (e.g. drums, chords, vocals) as input, and produces bass tracks both in audio and MIDI format.
It relies on a neural network architecture that models the relationship between bass lines and accompanying material through a 2-dimensional latent space, which can be used as a control parameter by the user to create multiple variations of bass lines for the same conditioning material. See \citep{grachten2020bassnet} for more details, or the \href{https://sonycslparis.github.io/bassnet/}{BassNet introduction web page} for a non-technical overview and demonstration video.

\subsection{Methods of Analysis}
\label{sec:analysis}
Our analysis consists of a study of production strategies used by Hyper Music to produce the audio material of the song, as well as the stylistic aspects of the produced material, in particular the tracks generated using BassNet.

Based on a log of the production process kept by CSL researchers and the Digital Audio Workstation session of the song produced by Hyper Music, we create workflow diagrams showing how the final song result is produced.
We use these diagrams to clarify the musical idiom related to BassNet outputs featured in the song.

To present and discuss the pitch content of the audio material, we make use of Fourier spectra/spectrograms, as well as automatic fundamental pitch detection using CREPE \citep{kim2018crepe}.
In our analysis we explicitly focus on the fact that when the timbre of an instrument is rich (either by itself or after effects processing), there is not always a one-to-one relation between nominal (notated) pitch and perceived pitch, due to the harmonic overtone structure.
To this end we perform manual transcription on some of the audio material.
In these cases we distinguish between two kinds of transcription.
The first kind is a transcription of the perceived pitches. We will refer to this kind of transcription as \emph{perceptual transcription}.
The second kind is a transcription of the notes that are perceived as being played.
We refer to this kind of transcription as \emph{monophonic transcription}.
In the latter kind the aim is to produce the notes that a bass player would play on their instrument to produce the audio material to be transcribed, whereas in the former kind the aim is to notate the pitch sensations that the audio material produced. 

Part of our analysis concerns pitch perception, which is hard to quantify.
We realize that this makes it especially important to provide ample evidence for the reader to corroborate by themselves.
important to provide ample evidence for the reader to corroborate.
To this end we provide a web page with supplementary figures and audio material at the following URL\footnote{In addition to the page hosted at the URL, the supplementary is published as \cite{maarten_grachten_2022_6896283}}:

\begin{center}
  \small
  \url{\smurl}
\end{center}

We strongly encourage the reader to consult the supplementary material, which we will reference frequently in the following sections.

\section{Results}\label{sec:results}
In this section, we present a selection of results from the collaborative session between Hyper Music and CSL.
Two songs were produced in this session, titled ``Melatonin'' and ``Skream'', but for brevity we will discuss only the former. 
We start with a brief overview of the production strategies employed by Hyper Music in the session, followed by a description of tweaks to the sonification process of the AI tool during the session, in response to suggestions by the artists.
After that we analyze the produced material in detail, highlighting style-related aspects.
 
\subsection{Production Strategies}
\label{sec:workflow}
Following \citep[p.5]{burgess2013art}, we refer to the term ``production'' as ``the technological extension of composition and orchestration.''
Production involves aspects of the music that can be transcribed as symbolic representation (e.g. as a score, ``composition''), as well as what can generally be considered as sound (``orchestration'').
In the BassNet session the technological extension of orchestration (mainly by way of BassNet and effects processing) is tightly interleaved with composition itself, in a process that is commonly labeled ``in-studio composition'' \citep{tamm1995brian,deruty2022development}.

With the term ``production strategies'', we refer to the ensemble of creative processes initiated and followed by the musicians when they build music content with studio equipment.
In the case of the BassNet session, studio equipment consists of the BassNet tool (in the form of a web interface to a GPU-powered server running the model), a Digital Audio Workstation with several commercial audio and instrument plug-ins, and other CSL prototypes (\href{\smurl#melatonin_luc}{Supplementary\ Material\ A.1}).

Hyper Music prepared some audio material ahead of the session, to serve as a basis for the song.
This material was used as conditioning material to produce outputs with BassNet (\href{\smurl#melatonin_wf_overall}{Suppl.\ Mat.\ A.2}).
Some of the BassNet outputs that were selected for inclusion in the song were subsequently used (along with the other conditioning audio) to produce further BassNet outputs.
Most of the outputs were processed using a variety of effects that are common in studio production of bass, guitar and other instruments nowadays, including software emulators of effect-pedals, 
amplifier, and speaker cabinets.

\subsection{Tweaks to BassNet sonification}
\label{sec:sound}
The audio output of BassNet is created using additive synthesis of sine waves, where at each frame the harmonic series is determined by BassNet's predicted fundamental frequency ($f_0$) for that frame, and the relative amplitude of the harmonics is inferred from the predicted CQT spectrum for that frame.
This---rather simplistic---sonification method has a tendency to sound synthetic an non-expressive, and was cited by Hyper Music as a limitation of the tool.
In response to suggestions by Hyper Music to create a richer sound, changes were made to the sonification method during the session, providing control over 1) portamento (continuous transitions between notes) 2) the relative strength of odd vs even harmonics, and 3) ``harmonic variation''.
We use the latter term to refer to a temporal envelope over the harmonics, achieved by a straight-forward non-linear transformation of the harmonic amplitudes from the CQT spectral frames (See \href{\smurl#harm_var}{Suppl.\ Mat.\ B}).
The transformation makes the output sound more sensitive to details of the conditioning material, and makes changes of timbre over the course of notes more salient.

The BassNet sounds used in ``Melatonin'' are shaped using these controls in accordance with Hyper Music's preferences: attenuating even harmonics, using slight portamento at times, and strong harmonic variation.

\subsection{Relations between Bass Lines -- Heterophony} \label{sub:relations_between_bass_lines}
\begin{figure}
  \centering
\includegraphics[width=\textwidth]{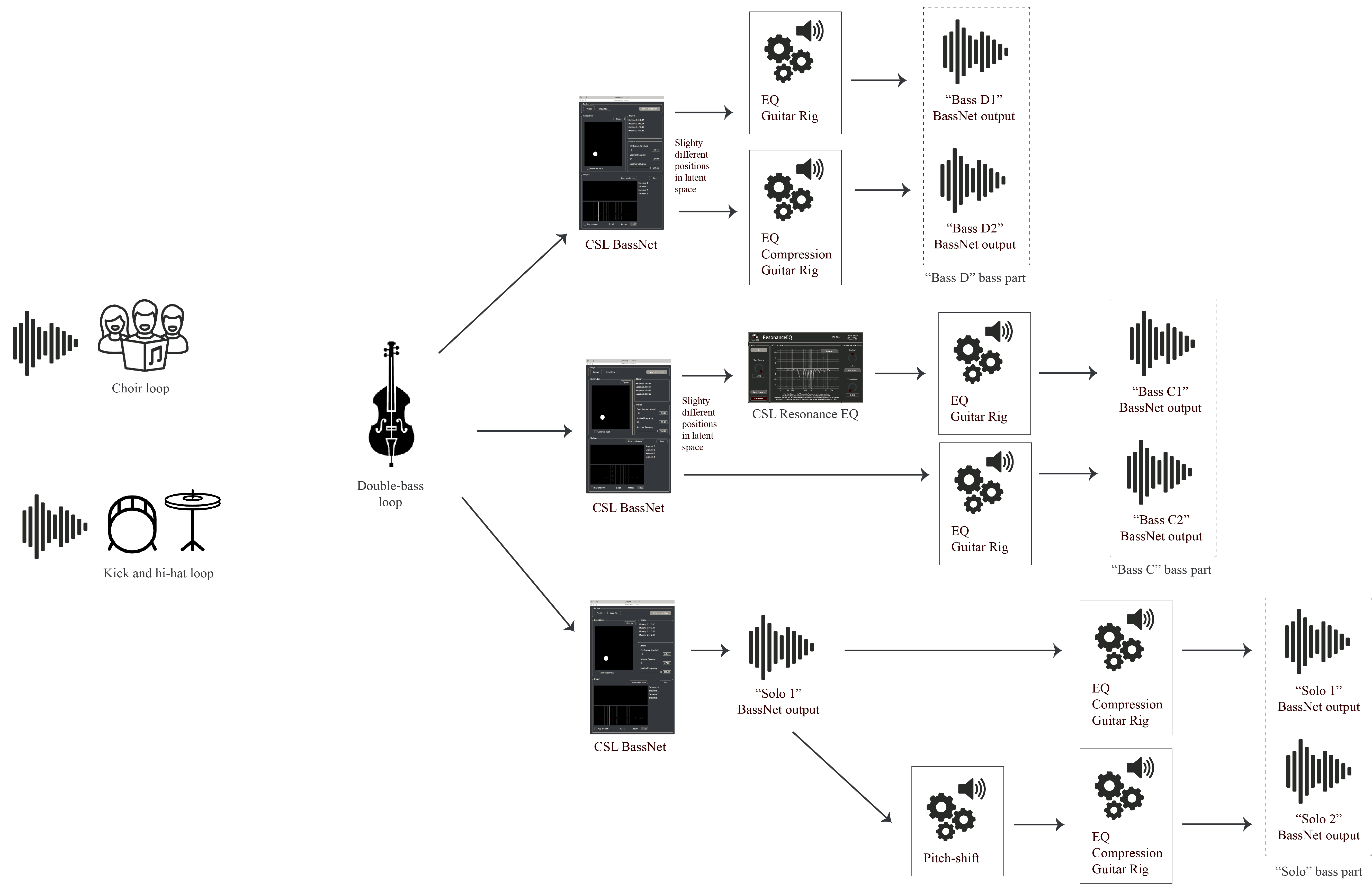}
\caption{Workflow for ``Melatonin'', Section 2.}
\label{fig:Workflow_Melatonin_2}
\end{figure}

The song (\href{\smurl#melatonin_song}{Suppl.\ Mat.\ A.3}) consists of two main sections, henceforth referred to as Section 1 and Section 2.
A production strategy used extensively in both sections was the layering of BassNet outputs.
The song features a total of 15 single BassNet outputs, grouped together to form composite parts. Figure~\ref{fig:Workflow_Melatonin_2} summarizes the creation/grouping of the individual parts for Section 2 
(For more detailed schemas of both sections with sound examples see \href{\smurl#melatonin_wf_1}{Suppl.\ Mat.\ A.4} and \href{\smurl#melatonin_wf_2}{Suppl.\ Mat.\ A.5}).

A production strategy used by Hyper Music in Section 1 was an iterative generation process, where BassNet outputs are conditioned on a mix of the original material plus BassNet outputs obtained earlier in the process (\href{\smurl#melatonin_wf_overall}{Suppl.\ Mat.\ A.2}).
In Section 1 (\href{\smurl#melatonin_wf_1}{Suppl.\ Mat.\ A.4}), the ``Bass B'' composite part was generated last, using the groups ``Bass A'', and ``High pattern'' as part of the conditioning.

Inside the composite parts, one relation between individual lines is doubling -- see \citep[p. 46]{moore2016song}.
Hyper Music used a pitch-shifter to add a second line to the original one. Another relation is timbre enrichment.
Hyper Music provided elements of one BassNet output to CSL prototype Notono \citep{bazin2021spectrogram}.
Using this tool BassNet outputs were transformed by means of analysis-resynthesis.
The transformed outputs are superimposed to the original output, thus providing additional timbre elements to the original line.

A third relation between individual lines in composite parts is \textit{heterophony}.
Heterophony is described as the occurrence of ``simultaneous variation of a single melody'' \citep{OUP_Heterophony}, and elsewhere as ``polyphony resulting from simultaneous differences of pitch produced when two or more people sing or play more or less the same melodic line at the same time.'' \citep[p. 210]{tagg2014everyday}.

\begin{figure}
  \centering
        \includegraphics[width=.85\textwidth]{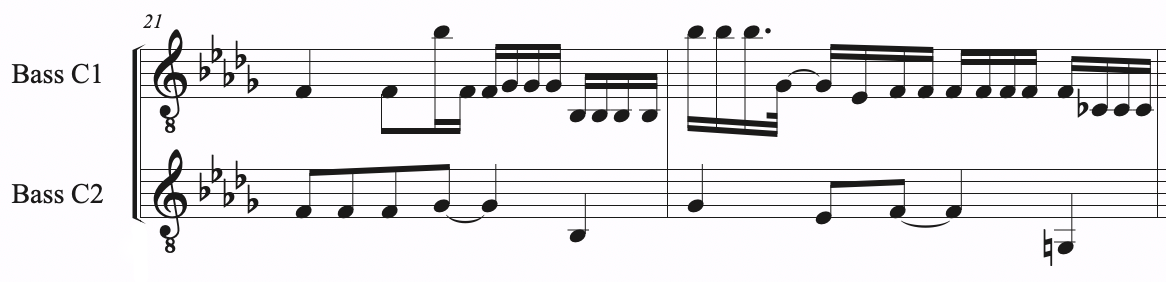}
        \includegraphics[width=.9\textwidth]{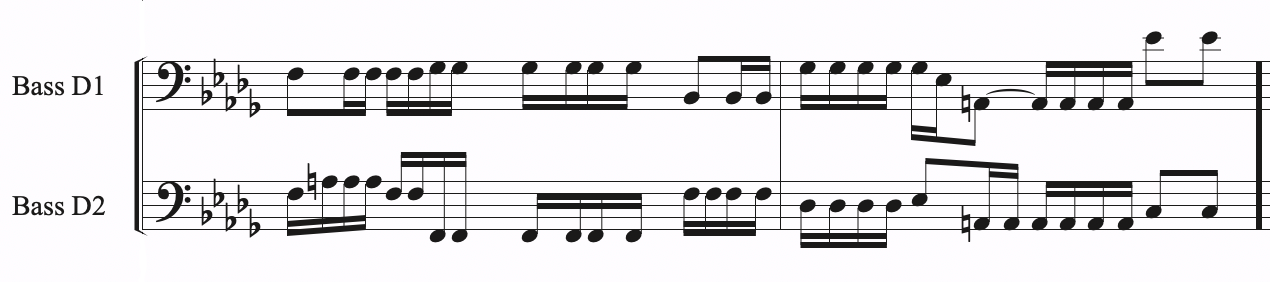}
        \caption{Two transcribed excerpts from Section 2 of ``Melatonin'' (monophonic transcription, see Section~\ref{sec:analysis}). Top: Bars 21 and 22 from parts ``Bass C1'' and ``Bass C2''. Bottom: Bars 27 and 28 from parts ``Bass D1'' and ``Bass D2''.
          Sound excerpts in \href{\smurl\#idiom_1_ex_1}{Suppl.\ Mat.\ C.1} and \href{\smurl\#idiom_1_ex_2}{Suppl.\ Mat.\ C.2}
        }
        \label{fig:BN_heterophony}
\end{figure}

BassNet can generate different suggestions based on the same audio conditioning.
Given a single conditioning, if two such suggestions derive from a close enough position in BassNet's latent space, then the two generated outputs can be similar to each other yet with clear differences. The superimposition of such outputs can produce heterophony.
Hyper Music used this production strategy throughout the song.
Two examples of this from Section 2 of the song are shown in Figure \ref{fig:BN_heterophony}.
The first example (top) contains two lines that are rhythmically distinct, but closely tied in terms of pitch, where ``Bass C1'' functions like an embellished variation of ``Bass C2'', and thus fits the definition of heterophony in the above sense.
By contrast, the parts ``Bass D1'' and ``Bass D2'' in the second example (bottom) are not obviously variations of a single melody in terms of pitch contour, and as such are not a typical example of heterophony.
Nevertheless, they are strongly tied rhythmically, sharing a tendency towards repeated 16\textsuperscript{th} notes.

Both sections of the song make heavy use of the heterophony idiom by superimposition of bass parts (as witnessed from the video material in \href{\smurl#melatonin_wf_1}{Suppl.\ Mat.\ A.4}, bottom, and \href{\smurl#melatonin_wf_1}{Suppl.\ Mat.\ A.5}, bottom).

\subsection{Homophony \emph{within} Bass Lines}\label{sec:homoph-emphw-bass}
In the Western classical music tradition, \textit{pitch} is ``[t]he particular quality of a sound (e.g. an individual musical note) that fixes its position in the scale'' \citep{Pitch}.
As such, if the material played by an instrument is monophonic (consists of a single voice or part), then a single note---and thus a single pitch---is expected to sound at any given time.

Even if bass parts are commonly conceived as monophonic---evident from the term ``bass line'', often used as a synonym for bass part---it is not always the case that they convey just a single pitch.
In contemporary Popular Music for example, bass parts may convey multiple simultaneous pitches. They are typically the result of post-hoc effects-processing or equalization of the original sound.
Three examples of this phenomenon from commercial mixes are discussed in detail in \href{\smurl#homophony_ex}{Suppl.\ Mat.\ D}. 

In Section \ref{sub:relations_between_bass_lines} we focused on simultaneous pitches (more specifically heterophony) as a result of the superimposition of related but different BassNet outputs.
An examination of individual outputs reveals that simultaneous pitches even occur within a single BassNet output.

\begin{figure}
  \centering
  \includegraphics[width=.9\textwidth]{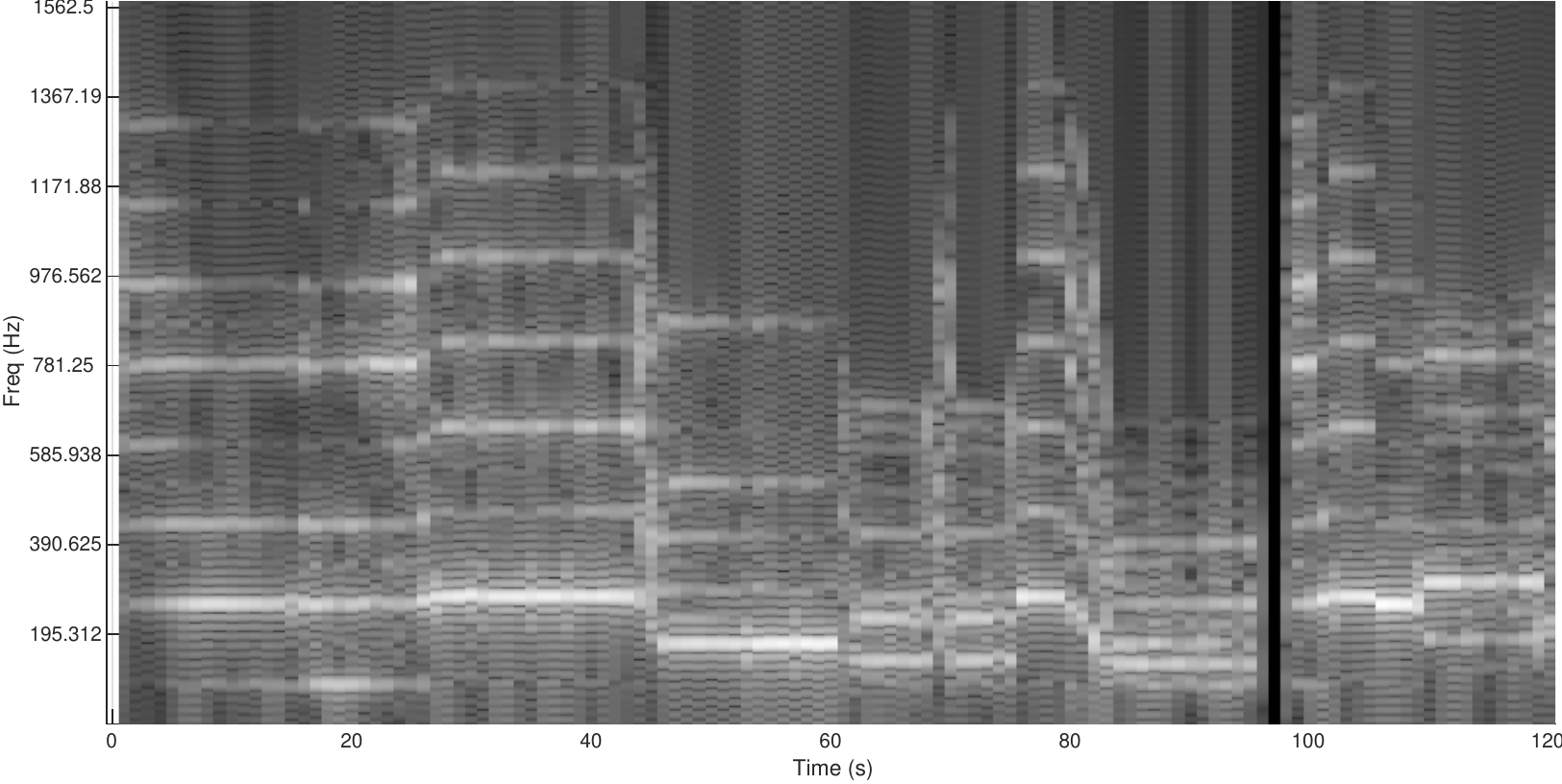}
  \includegraphics[width=.7\textwidth]{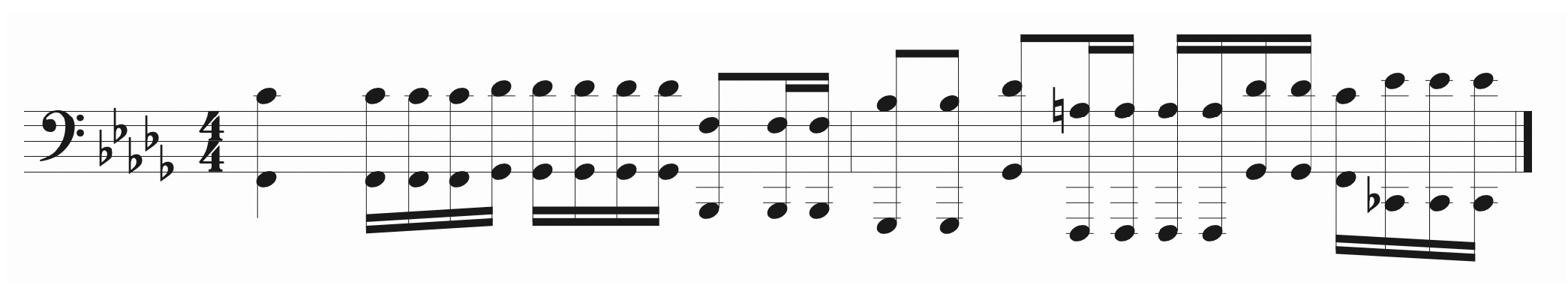}
  \caption{Melatonin, section 2, track ``Solo 1'', bars 82--83. Top: Fourier spectrogram; Bottom: Perceptual transcription (see Section~\ref{sec:analysis}). Corresponding sound excerpt in \href{\smurl\#solo1_analysis}{Suppl.\ Mat.\ E.1}}
  \label{fig:Melatonin2_lines}
\end{figure}

As an example of this, Figure \ref{fig:Melatonin2_lines} shows the Fourier spectrogram of Section 2, Solo 1, bars 82--83, along with a perceptual transcription of the part.
The transcription was performed by a combination of listening and inspection of Fourier spectra.
To this end the spectra of selected frames (computed with higher frequency resolution) are plotted and annotated, as shown in \href{\smurl#solo1_analysis}{Suppl.\ Mat.\ E.1}.

The transcription conveys two simultaneous pitches throughout the part, while the interval between simultaneous pitches varies over time.
The associated spectra (\href{\smurl#solo1_analysis}{Suppl.\ Mat.\ E.1}) generally show attenuated even harmonics, a consequence of the artists' preference of odd over even harmonic sonification in BassNet.
Furthermore, the spectra reveal that the temporal variability in the amplitude of the harmonics, including attenuation of the two lowest harmonics (as a result of the ``harmonic variation'' control, see Section~\ref{sec:sound} and \href{\smurl#harm_var}{Suppl.\ Mat.\ B}) translates into a variety of modalities for the perception of pitch.
The part has a permanent audible second line, which often follows harmonic 3 but not all the time.

Similar to the bass line examples from well-known commercial mixes mentioned above (\href{\smurl#heterophony_ex}{Suppl.\ Mat.\ D}), individual BassNet outputs are close to being polyphonic, where the simultaneous progressions of pitch are tied rhythmically.
In this respect it resembles homorhythmic homophony \citep{Homophony}, even if the term homophony (just like polyphony in general) is typically used to denote music where each pitch is produced by a separate note.

\section{Discussion}\label{sec:discussion}
The conditioning material used by Hyper Music consists of various elements, including voice, chords, and drums.
For Section 2 it consists of a simple acoustic double bass part.
The conditioning bass part constrains the variability of BassNet outputs as a function of the latent space controls, such that the outputs tend to remain close to the conditioning part.
In this respect Hyper Music used BassNet not so much to come up with a primary idea for Section 2, as to develop a single part into a Section.
The development in this section consists in a time-varying superimposition of multiple variations of BassNet outputs, forming rich heterophonic textures (see Section~\ref{sub:relations_between_bass_lines}).

Using \cite{OUP_Style}'s terms (Section~\ref{sec:background}), such instances of heterophony may be considered as an idiom derived from the resources of the performance.
The individual parts that make up the heterophonic texture were created using an intrinsic affordance of BassNet: its latent space control makes it straight-forward to create different voices fitting the same conditioning material.

In addition, the individual audio outputs of BassNet also introduce a notable style, namely the occurrence of simultaneous pitches in a monophonic bass line (see Section~\ref{sec:homoph-emphw-bass}).
This style can also be found in various existing commercial mixes, where it is typically achieved by production techniques like effects processing, double tracking \citep[p. 46]{moore2016song}, and equalization.
In contrast to those techniques, the overtone structure of BassNet outputs is directly produced as a function of the conditioning audio, inferring harmonic amplitudes from the predicted CQT spectra.

The permanence of a second perceived pitch value in some outputs is reminiscent of overtone singing, in which a single vocal performer produces more than one clearly audible pitch simultaneously \citep{Overtonesinging}.
Although superficially this seems to be a manifestation of style as characteristic \emph{sound}, we have highlighted how the near-polyphonic quality of the sound constitutes homorhythmic homophony~\citep{Homophony}.
As such it can also be regarded as a musical \emph{idiom}.
Note how this blurs the distinction we adopted from \cite{OUP_Style} in Section \ref{sec:background} where manifestations of style concern either the sound, or the idiom.

From the example presented in Section~\ref{sec:homoph-emphw-bass} we identified the ``odd vs even harmonic'' and the ``harmonic variation'' controls (changes to the BassNet sonification method prompted by suggestions from the artists, see Section~\ref{sec:sound}) as the source of the latter stylistic characteristics.
In particular we found that the absence of even harmonics, the attenuation of the first two harmonics and the temporal fluctuation of relative harmonic strengths leads to ambiguities in pitch perception.
The combination of the first two factors produces a scenario where the lowest component in the tone complex is an odd harmonic of the original $f_0$ and the spacing between harmonics (due to the attenuation of even harmonics) is $2f_0$.
This scenario has been shown to produce two pitch sensations---a phenomenon that poses problems for both spectral and temporal theories of pitch \citep[pp. 1705--1706]{yost2009pitch}.
Given Hyper Music's concern that the BassNet sonification before the adaptations tended to produce uninteresting sound qualities, it is conceivable that their preference for this particular setting is precisely because of the ambiguity of pitch perception it introduces.

Beyond observations about stylistic aspects of the output, this case study illustrates some notable points about the use of novel technology for creative musical purposes in general.
First, the additional controls to shape the way BassNet predictions are sonified were suggested by Hyper Music with an explicit reference to analog synthesizers as an effective way to interact with the technology.
There are multiple ways to interpret this reference.
On a basic level it may demonstrate the need of music artists to have fine-grained control of the sounds they produce.
But it may also serve as an illustration that in order for music artists to accept a tool and be able to interact with it intuitively, its interface must appeal to notions that are familiar to them. 
The history of the analog synthesizer itself also holds evidence for this.
For example, the commercial success of the Moog synthesizer---compared to another type of analog synthesizer introduced around the same time by Don Buchla---is explained by \cite{Pinch2002} as a result of Bob Moog's embrace of the familiar piano keyboard as a way to interact with his synthesizer (as opposed to Buchla, who rejected it on the grounds that it was too limited for the possibilities of the synthesizer).
In the same vein, collaborations with musician Herbert Deutsch led Moog to include time-varying filters that emulate the articulation that is characteristic of conventional acoustical instruments \citep[page 27]{Pinch2002}.

Second, it can be argued that the non-linear transformation of relative harmonic strengths of the sound (``harmonic variation'') that turned out to make the BassNet sound richer and more ambiguous in terms of pitch, is essentially a rather ad-hoc fix to the synthetic quality of the original sounds, because the BassNet model was not designed or trained to take this transformation into account.
As such, it is hard to see how this example informs the development of AI tools for music from a technical perspective.
Nevertheless it would be wrong to negate the role of such serendipity in musical creativity.
This role is clearly illustrated by the works of \cite{Schaeffer52} who recognized, for example, that something as simple as playing back a 78 RPM recording at 33 RPM can reveal a whole variety of musical forms in the modified audio content that were not apparent in the original, but may be the basis of a coherent musical discourse.
Therefore, we subscribe to the view---propagated among others by \cite{doi:10.1080/09298215.2018.1515233,deruty2022development}---that AI tools for music creativity should be developed alongside, and in collaboration with musicians.

\section{Conclusions}\label{sec:conclusions}
In this case study we have investigated style characteristics of a song by artist duo Hyper Music in relation to BassNet, an AI tool to produce bass lines that was used extensively in the music production process.
For this, we draw on the definition of style as either sound or musical idiom emerging from the resources of the performance, in this case BassNet specifically \citep{OUP_Style}.

One type of style that is prominent in the song is a heterophony \emph{idiom}, produced by superimposing variations of bass lines.
The variations are created by using the same conditioning material to BassNet with different configurations of the tool, notably different latent space positions \citep{grachten2020bassnet}.

During the collaborative session between Hyper Music and CSL in which the song was produced, tweaks were made to the sonification process of BassNet that resulted in richer and more time-varying timbres.
Spectral and perceptual analysis of BassNet outputs indicates that the resulting harmonic overtone structure often produces the sensation of two simultaneous pitches originating from a single note.
Although this phenomenon can be regarded as a manifestation of style through \emph{sound}, the parallel melodic lines that form due to the dual pitch percepts effectively form a homorhythmic homophony musical \emph{idiom} within a single instrument part.
This idiom also occurs in the bass parts of some commercial contemporary Popular Music---where the effect is typically produced by double tracking, equalization or other production strategies---and in overtone singing.

The case study presented here illustrates how AI tools used for music production---like any other technology or instrument---create affordances that, depending on the role they play in the music production process, may imbue musical outputs with a distinct style, both in terms of sound and musical idiom.
Moreover, it shows that bringing AI tools into music practice creates opportunities for improvements that make the tools more engaging for artists to work with.

\newpage 
\small
\bibliographystyle{apalike}

\bibliography{references.bib}

\end{document}